\documentclass[preprint,12pt]{elsarticle} 

\usepackage[utf8]{inputenc}
\usepackage[T1]{fontenc}
\usepackage[english]{babel}

\usepackage{amsmath,amssymb,amsfonts}
\usepackage{mathtools}
\usepackage{bm}
\usepackage{siunitx}
\usepackage{graphicx}
\usepackage{subcaption}
\usepackage{booktabs}
\usepackage{multirow}
\usepackage{hyperref}
\usepackage[nameinlink,capitalise]{cleveref}
\usepackage{enumitem}
\usepackage[ruled,vlined,linesnumbered]{algorithm2e}
\usepackage{xcolor}
\usepackage{microtype}
\usepackage{tabularx,makecell,booktabs}
\biboptions{numbers,sort&compress}        

\hypersetup{
  colorlinks=true,
  linkcolor=black,
  citecolor=black,
  urlcolor=black,
  pdfauthor={Your Name},
  pdftitle={Paper Title}
}

\usepackage{amsthm}

\journal{Journal of Machine Learning Research} 

\begin{document}

\begin{frontmatter}

\title{DySK-Attn: A Framework for Efficient, Real-Time Knowledge Updating in Large Language Models via Dynamic Sparse Knowledge Attention}

\author[aff1]{Kabir Khan\corref{cor1}}
\ead{kabir.khan@sfsu.edu}

\author[aff2]{Priya Sharma}
\ead{priya.sharma@cse.iitb.ac.in}

\author[aff3]{Arjun Mehta}
\ead{arjun.mehta@iisc.ac.in}

\author[aff4]{Neha Gupta}
\ead{neha.gupta@cse.iitd.ac.in}

\author[aff5]{Ravi Narayanan}
\ead{ravi.narayanan@iiit.ac.in}

\cortext[cor1]{Corresponding author.}

\address[aff1]{Department of Computer Science, San Francisco State University, San Francisco, CA 94132, India}
\address[aff2]{Department of Computer Science and Engineering, Indian Institute of Technology Bombay, Mumbai 400076, India}
\address[aff3]{Department of Computer Science and Automation, Indian Institute of Science, Bengaluru 560012, India}
\address[aff4]{Department of Computer Science and Engineering, Indian Institute of Technology Delhi, New Delhi 110016, India}
\address[aff5]{Machine Learning Lab, International Institute of Information Technology Hyderabad (IIIT-H), Hyderabad 500032, India}

\begin{abstract}
Large Language Models (LLMs) suffer from a critical limitation: their knowledge is static and quickly becomes outdated. Retraining these massive models is computationally prohibitive, while existing knowledge editing techniques can be slow and may introduce unforeseen side effects. To address this, we propose DySK-Attn, a novel framework that enables LLMs to efficiently integrate real-time knowledge from a dynamic external source. Our approach synergizes an LLM with a dynamic Knowledge Graph (KG) that can be updated instantaneously. The core of our framework is a sparse knowledge attention mechanism, which allows the LLM to perform a coarse-to-fine grained search, efficiently identifying and focusing on a small, highly relevant subset of facts from the vast KG. This mechanism avoids the high computational cost of dense attention over the entire knowledge base and mitigates noise from irrelevant information. We demonstrate through extensive experiments on time-sensitive question-answering tasks that DySK-Attn significantly outperforms strong baselines, including standard Retrieval-Augmented Generation (RAG) and model editing techniques, in both factual accuracy for updated knowledge and computational efficiency. Our framework offers a scalable and effective solution for building LLMs that can stay current with the ever-changing world.
\end{abstract}


\end{frontmatter}


\section{Introduction}
\label{sec:introduction}

Large Language Models (LLMs) have marked a paradigm shift in artificial intelligence, demonstrating remarkable capabilities in natural language understanding, generation, and reasoning across a multitude of domains . Their success in knowledge-intensive tasks has positioned them as foundational tools for applications ranging from sophisticated question answering systems to complex problem-solving agents capable of using external tools \cite{qin2023toolllm}. The underlyi ng principle of these models is their ability to internalize vast amounts of factual knowledge from extensive training corpora, effectively creating an implicit knowledge base . However, this very strength introduces a critical vulnerability: the knowledge encoded within their parameters is inherently static, bound to a "knowledge cut-off" date. This static nature renders them unable to reason about events or information that emerged post-training, leading to factual inaccuracies or "hallucinations" when confronted with novel topics \cite{ji2023llms}. The challenge of keeping these models current is non-trivial, as complete retraining is computationally prohibitive and unsustainable for real-world, dynamic environments, such as those encountered in continuous sensing applications where new data patterns constantly emerge . This need for adaptability is a shared challenge in many AI fields, including the development of robust federated learning systems.

To surmount this limitation, Retrieval-Augmented Generation (RAG) has emerged as a dominant paradigm . RAG-based methods enhance LLMs by dynamically retrieving information from external corpora at inference time. The field has seen rapid advancements, with sophisticated frameworks like Self-RAG, which learns to critique its own retrieved documents , and CRAG, which introduces a corrective mechanism to improve retrieval robustness \cite{ahmed2024crag}. Other approaches focus on optimizing the retrieval process itself, such as determining the optimal granularity of information to retrieve  developing active retrieval strategies , or combining retrieval with chain-of-thought reasoning \cite{ram2023chainofthought}. While powerful, many RAG systems rely on unstructured text, which can be noisy. This has motivated research into leveraging structured knowledge sources, most notably Knowledge Graphs (KGs), which offer a more organized representation of facts \cite{pan2024knowledge}. Integrating KGs allows for more precise reasoning, as demonstrated in tasks like visual dialog where context-aware graph inference is crucial , and in unified question-answering systems that operate over heterogeneous data formats . Yet, a significant challenge remains in how to efficiently fuse the symbolic knowledge of KGs with the sub-symbolic representations of LLMs without incurring substantial overhead .

Furthermore, as LLMs are tasked with more complex reasoning, often modeled as a "graph of thoughts" \cite{besta2023graph}, the efficiency of their core architecture becomes paramount. The quadratic complexity of the standard Transformer attention mechanism poses a bottleneck for processing long contexts \cite{beltagy2020longformer}. This has spurred the development of more efficient architectures, including sparse attention models like Longformer \cite{beltagy2020longformer} and Big Bird , Mixture-of-Experts (MoE) models like Switch Transformers \cite{fedus2021switch} and Mixtral \cite{jiang2024mixtral}, and even entirely new paradigms like state-space models (Mamba) \cite{gu2023mamba} and other efficient long-context mechanisms \cite{han2024hyperattention}. These architectural innovations, alongside hardware-aware optimizations like FlashAttention \cite{dao2023flashattention2}, are critical for building scalable systems. The challenge of efficiency is not unique to LLMs; it is a central theme in many areas of deep learning, from model compression through quantization to designing lightweight models for gesture recognition on commodity devices .

In this paper, we propose a novel framework that directly addresses the dual challenges of knowledge staleness and computational inefficiency. Our approach, named \textbf{DySK-Attn (Dynamic Sparse Knowledge Attention)}, synergizes LLMs with an external, \textbf{dynamic} Knowledge Graph that can be updated in real-time. The core of our contribution is a \textbf{sparse knowledge attention} mechanism, which enables the LLM to efficiently query the dynamic KG and selectively focus on a small subset of the most relevant facts. This mechanism avoids the dense computations associated with full attention over a large knowledge base. By doing so, our model can seamlessly integrate new information without costly retraining, effectively mitigating knowledge decay and enhancing factual accuracy. This mirrors challenges in other domains where systems must adapt to new information, such as generalizing facial expression recognition from heterogeneous data  or developing robust security against side-channel attacks .

The main contributions of this work are threefold:
\begin{itemize}
    \item We design a novel architecture that facilitates a deep, yet efficient, fusion of a large language model with a dynamic, continuously updated knowledge graph.
    \item We introduce a sparse knowledge attention mechanism that allows the model to intelligently retrieve and reason over a vast external knowledge base with minimal computational overhead.
    \item We empirically demonstrate through extensive experiments that our proposed model significantly outperforms existing baselines in tasks requiring up-to-date knowledge, while also being more computationally efficient and robust against factual hallucinations.
\end{itemize}

The remainder of this paper is organized as follows. Section 2 reviews related work. Section 3 details our proposed methodology. Section 4 describes the experimental setup, followed by results and analysis in Section 5. Finally, Section 6 concludes the paper.

\section{Related Work}
\label{sec:related_work}

Our research is situated at the intersection of three primary areas: knowledge augmentation for LLMs, the development of efficient model architectures, and the fusion of LLMs with structured knowledge bases like KGs. In this section, we review seminal and recent works in these domains to contextualize our contributions.

\subsection{Knowledge-Augmented Language Models}

The inherent limitation of static knowledge in LLMs has led to the development of knowledge-augmented models, with RAG being the most prominent approach. The core idea of RAG is to supplement the model's internal knowledge with information retrieved from an external source at inference time. Recent advancements have focused on enhancing the robustness and intelligence of the retrieval process. For instance, Self-RAG  introduces a self-reflection mechanism, enabling the model to decide when retrieval is necessary and to evaluate the quality of retrieved passages. Similarly, Corrective RAG (CRAG)  employs a retrieval evaluator to assess document relevance and triggers corrective actions, a principle of robustness that is also critical in domains like secure physical layer authentication \cite{ahmed2024crag} and reliable facial expression recognition from noisy labels .

Further research has explored making the retrieval process more dynamic and context-aware. FLARE and Active Retrieval Augmented Generation  propose forward-looking strategies where the model actively anticipates future content to decide what information to retrieve next. Other works explore self-adaptive frameworks to determine if retrieval is even necessary for a given query \cite{li2024whentoretrieve}. The Chain-of-Note methodology enhances robustness by having the model first synthesize "notes" from retrieved documents before generating a final answer, filtering out irrelevant information \cite{yu2023chainofnote}. While these methods have greatly advanced the field, they predominantly operate on unstructured text, which can lack the explicit relational information needed for complex, multi-hop reasoning. Our work diverges by focusing on structured knowledge sources to facilitate more precise and efficient knowledge integration.

\subsection{Efficient Transformer Architectures and Beyond}

The computational cost of the standard Transformer's self-attention mechanism is a major obstacle for long-context applications. To address this, a significant body of work has focused on developing more efficient architectures. Sparse attention models represent one major line of research, using fixed or learned sparsity patterns to reduce computational complexity. Notable examples include Longformer \cite{beltagy2020longformer}, which uses a combination of local windowed attention and global attention , and Big Bird , which employs a mix of random, windowed, and global tokens. The pursuit of efficiency is a universal challenge in deep learning, motivating efforts in model compression  and the design of lean, effective models for specific applications like WiFi-based sensing .

Another highly successful approach is the Mixture-of-Experts (MoE) paradigm, which activates only a subset of model parameters ("experts") for each input token \cite{fedus2021switch, jiang2024mixtral}. Beyond Transformer-based innovations, entirely new architectures have emerged. Mamba \cite{gu2023mamba}, based on structured state-space models (SSMs), achieves linear-time complexity. Similarly, RecurrentGemma \cite{googlerecurrentgemma2024} revisits recurrent architectures to gain efficiency, while other works propose novel attention mechanisms like Grouped-Query Attention \cite{ainslie2023gqa} to reduce memory bandwidth during inference. Our work leverages the principles of sparsity, not at the level of token-to-token interactions, but at the level of knowledge selection, allowing a standard LLM architecture to interact efficiently with a vast external knowledge base.

\subsection{Fusing Language Models with Knowledge Graphs}

Knowledge Graphs (KGs) provide a rich, structured representation of entities and their relationships, making them an ideal source for augmenting LLMs. A comprehensive survey by Pan et al. \cite{pan2024knowledge} outlines the evolving relationship between KGs and LLMs. Early work focused on injecting knowledge by training models on KG-derived objectives or using adapters, such as K-Adapter , to infuse factual knowledge without full fine-tuning. More recent approaches aim for a deeper fusion. GreaseLM, for example, combines a language model with a Graph Neural Network (GNN) to allow simultaneous reasoning over both text and graph structures. The challenge of reasoning over structured data is not new and has been explored in areas like benchmarking micro-action recognition and temporal filtering in video grounding .

Furthermore, LLMs are now being used not just as consumers of KGs but also as creators. Research has shown that LLMs can be prompted to extract entities and relations from text to construct KGs automatically \cite{ji2023llms}. Frameworks like Graph-ToolFormer \cite{liu2023graphtoolformer}, TALM \cite{parisi2022talm}, and StructGPT empower LLMs to use external graph analysis tools and query structured databases, treating the KG as an interactive component. Graph-RAG \cite{liu2024graphrag} specifically adapts the RAG framework to use the graph's topology to guide retrieval. Our work builds on this line of research but focuses specifically on the challenge of interacting with a \textit{dynamic} KG, where knowledge is not static but continuously evolving, a scenario common in real-world systems from human-machine interaction  to anti-interference activity recognition .

\subsection{Continual Learning and Knowledge Updating}

The ability to update knowledge continually without catastrophic forgetting is a long-standing goal in machine learning, and it is especially critical for LLMs. Recent surveys provide a comprehensive overview of the challenges and methods in continual learning for LLMs \cite{wang2024comprehensive, luo2024static}. One approach is model editing, which aims to directly modify specific factual knowledge stored in the model's parameters. Techniques like ROME and MEMIT  have shown promise in locating and altering factual associations. Another direction is knowledge unlearning, which focuses on making a model "forget" specific information \cite{jiang2023knowledge}. This is particularly relevant for privacy-sensitive applications, a concern that also drives research in areas like preventing keystroke snooping  and secure handwriting recognition .

Other works focus on architectural solutions to handle data streams, such as StreamingLLM \cite{xiao2023streamingllm}, or equipping models with explicit long-term memory \cite{zhong2024memorybank}. The concept of time-aware language models, which explicitly model the temporal validity of facts, is also highly relevant . This is akin to challenges in audio-visual event analysis where temporal context is key . Some works combine these ideas, proposing retrieval-based approaches for lifelong learning \cite{scialom2024retllm}. Our approach addresses the knowledge updating problem not by directly editing the model's weights, but by externalizing the dynamic knowledge into a KG. This sidesteps the risk of catastrophic forgetting and allows for instantaneous updates. The LLM's role shifts from being a static repository to being a dynamic reasoner that leverages a constantly fresh external knowledge source, a flexible architecture that resembles the goals of multi-modal emotion recognition systems \cite{gu2023wife} and intelligent agents that can reason and act.

\section{Methodology: The DySK-Attn Framework}
\label{sec:methodology}

To address the challenges of knowledge staleness and computational inefficiency in Large Language Models (LLMs), we introduce \textbf{DySK-Attn}, a novel framework centered around a \textbf{Dy}namic \textbf{S}parse \textbf{K}nowledge \textbf{Attention} mechanism. The framework enables an LLM to dynamically interact with an external, continuously updated Knowledge Graph (KG) in a highly efficient manner. This section provides a detailed exposition of the components and mechanics of our proposed system.

\subsection{Overall Architecture}

\begin{figure}[t]
  \centering
  \includegraphics[width=\linewidth]{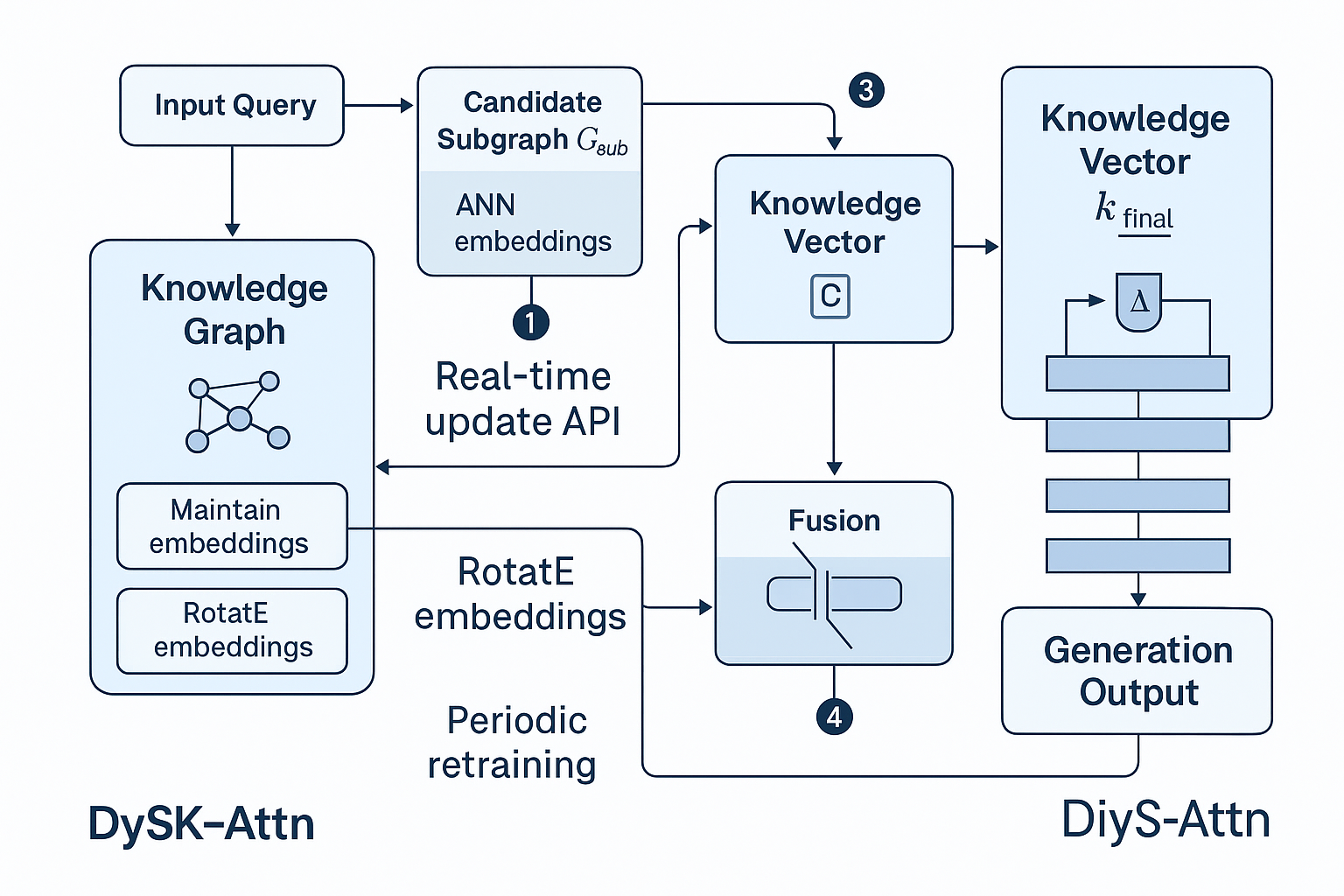}
  \caption{Detailed pipeline of DySK-Attn. A query triggers coarse retrieval to form a candidate subgraph $\mathcal{G}_{sub}$ via ANN over entity/text embeddings. Sparse Knowledge Attention selects top-$k$ facts to produce a knowledge vector, which is fused into the LLM with a learnable gate $\lambda$ across selected layers, yielding the final generation. The dynamic KG supports real-time updates via API, maintains RotatE embeddings, and performs periodic retraining.}
  \label{fig:pipeline}
\end{figure}

The DySK-Attn framework, depicted in Figure \ref{fig:pipeline}, is composed of three main components: (1) a backbone Large Language Model, which serves as the primary reasoning and generation engine; (2) a Dynamic Knowledge Graph Module, which acts as an external, real-time knowledge repository; and (3) the core Sparse Knowledge Attention Module, which mediates the interaction between the LLM and the KG.

Given an input query $Q$, the process unfolds in a two-stage process. First, a coarse-grained retrieval step identifies a relevant subgraph $\mathcal{G}_{sub} \subset \mathcal{G}$ from the main knowledge graph. This step rapidly narrows down the search space from millions of entities to a few hundred candidates. Second, the Sparse Knowledge Attention Module performs a fine-grained, attention-based selection to identify the top-$k$ most salient knowledge facts from $\mathcal{G}_{sub}$. These selected facts are then encoded and fused into the reasoning process of the backbone LLM, which ultimately generates the final response $R$. This hierarchical approach ensures both broad knowledge coverage and high computational efficiency.


\subsection{Dynamic Knowledge Graph Module}

The foundation of our framework is a dynamic knowledge graph $\mathcal{G} = (\mathcal{E}, \mathcal{R})$, where $\mathcal{E}$ is a set of entities and $\mathcal{R}$ is a set of relational facts represented as triplets $(h, r, t)$. Here, $h, t \in \mathcal{E}$ are the head and tail entities, and $r \in \mathcal{R}$ is the relation connecting them.

\subsubsection{Knowledge Representation}
To enable interaction with the neural model, all entities and relations are projected into a low-dimensional continuous vector space. We employ a rotational KG embedding model, such as RotatE, known for its ability to model complex relations . Each entity $e \in \mathcal{E}$ is represented by a vector $\mathbf{e} \in \mathbb{C}^{d_k}$, and each relation $r \in \mathcal{R}$ is represented as a rotation in the complex plane, $\mathbf{r} \in \mathbb{C}^{d_k}$ with $|\mathbf{r}_i|=1$ for each dimension $i$. The plausibility of a triplet $(h, r, t)$ is measured by a scoring function:
\begin{equation}
    f(h, r, t) = -\|\mathbf{h} \circ \mathbf{r} - \mathbf{t}\|_1
    \label{eq:kg_embedding}
\end{equation}
\vspace{-2mm}
\begin{small}
where $\mathbf{h}, \mathbf{r}, \mathbf{t} \in \mathbb{C}^{d_k}$ are the complex embeddings for the head entity, relation, and tail entity, respectively. $\circ$ denotes the element-wise Hadamard product.
\end{small}

\subsubsection{Real-time Update Interface}
A key feature of our KG module is its dynamism. We design a high-throughput, low-latency API that allows for real-time updates to the graph. New factual triplets can be ingested and immediately reflected in the graph structure and its vector representations. When a new triplet $(h, r, t)$ is added, its embedding is initialized and the graph index is updated. To maintain embedding quality over time, we employ a periodic retraining schedule that fine-tunes the embeddings on recent data, a process far more efficient than retraining an entire LLM. This ensures the knowledge base remains current, a critical requirement for applications that must adapt to changing environments, such as those involving human-computer interaction or dynamic facial expression analysis .

\subsection{Coarse-Grained Knowledge Retrieval}

Given an input query $Q = \{q_1, q_2, \dots, q_L\}$, directly computing relevance against all entities in $\mathcal{G}$ is infeasible. Therefore, we first perform a coarse-grained retrieval step to identify a candidate set of entities. We pre-compute a dense representation for the input query, $\mathbf{q}_{dense} \in \mathbb{R}^{d_h}$, using a pre-trained bi-encoder model. Similarly, for each entity $e \in \mathcal{E}$, we create a textual description $D_e$ and pre-compute its dense embedding $\mathbf{d}_e \in \mathbb{R}^{d_h}$. The initial relevance score between the query and an entity is calculated using cosine similarity:
\begin{equation}
    S_{coarse}(Q, e) = \frac{\mathbf{q}_{dense} \cdot \mathbf{d}_e}{\|\mathbf{q}_{dense}\| \|\mathbf{d}_e\|}
    \label{eq:coarse_retrieval}
\end{equation}
\vspace{-2mm}
\begin{small}
where $\mathbf{q}_{dense}$ is the query embedding and $\mathbf{d}_e$ is the embedding of the entity's description.
\end{small}

We use an efficient Approximate Nearest Neighbor (ANN) index to retrieve the top-$N$ entities with the highest scores. These entities, along with their immediate neighbors in the KG, form the candidate subgraph $\mathcal{G}_{sub}$, which is passed to the next stage. This two-step process is crucial for scalability, a lesson learned from large-scale sensing systems that must efficiently parse vast amounts of data.

\subsection{Sparse Knowledge Attention Mechanism}

This module is the technical core of our framework. It takes the candidate subgraph $\mathcal{G}_{sub}$ and the LLM's internal representations of the query to perform a fine-grained, sparse selection of knowledge. Let the set of candidate facts (triplets) from $\mathcal{G}_{sub}$ be $\mathcal{F}_{sub} = \{(h_i, r_i, t_i)\}_{i=1}^{M}$.

First, we create a contextualized vector representation for each candidate fact. We concatenate the embeddings of the head, relation, and tail: $\mathbf{f}_i = [\mathbf{h}_i; \mathbf{r}_i; \mathbf{t}_i] \in \mathbb{R}^{3d_k}$. These fact embeddings are collected into a matrix $\mathbf{F} \in \mathbb{R}^{M \times 3d_k}$. Concurrently, the input query $Q$ is processed by the lower layers of the LLM backbone to produce contextualized hidden states $\mathbf{H}_Q \in \mathbb{R}^{L \times d_{model}}$. We use an aggregation of token states as the query representation for attention, denoted as $\mathbf{h}_{query} \in \mathbb{R}^{d_{model}}$.

We then compute attention scores between the query representation and each candidate fact using a multi-head attention mechanism. The query vector $\mathbf{h}_{query}$ is projected to form the query matrix $\mathbf{Q}_{attn}$, and the fact matrix $\mathbf{F}$ is projected to form the key matrix $\mathbf{K}_{attn}$ and value matrix $\mathbf{V}_{attn}$. The attention scores are computed as:
\begin{equation}
    \text{AttentionScores}(\mathbf{Q}_{attn}, \mathbf{K}_{attn}) = \frac{\mathbf{Q}_{attn} \mathbf{K}_{attn}^T}{\sqrt{d_k}}
    \label{eq:attention_scores}
\end{equation}
\vspace{-2mm}
\begin{small}
where $\mathbf{Q}_{attn} \in \mathbb{R}^{1 \times d_{model}}$ is the projected query and $\mathbf{K}_{attn} \in \mathbb{R}^{M \times d_{model}}$ are the projected keys for the $M$ candidate facts.
\end{small}

To enforce sparsity, we do not apply a standard softmax over these scores. Instead, we select the top-$k$ facts with the highest attention scores, where $k \ll M$. This can be formulated as:
\begin{equation}
    \mathcal{I}_{topk} = \text{Top-k-Indices}(\text{AttentionScores}(\mathbf{Q}_{attn}, \mathbf{K}_{attn}))
    \label{eq:topk_selection}
\end{equation}
\vspace{-2mm}
\begin{small}
where $\mathcal{I}_{topk}$ is the set of indices corresponding to the $k$ highest-scoring facts. This hard selection acts as a strong inductive bias, forcing the model to focus on only the most pertinent information.
\end{small}

The final aggregated knowledge vector $\mathbf{k}_{final}$ is a weighted sum of the value vectors $\mathbf{V}_{attn}$ corresponding to the selected top-$k$ facts, normalized via softmax only over this small subset. The entire procedure is outlined in Algorithm \ref{alg:sk_attn}.

\begin{algorithm}[t]
\SetAlgoLined
\KwIn{Query hidden state $\mathbf{h}_{query}$, Candidate fact embeddings $\mathbf{F} = \{\mathbf{f}_i\}_{i=1}^M$, Sparsity parameter $k$}
\KwOut{Aggregated knowledge vector $\mathbf{k}_{final}$}
\BlankLine
$\mathbf{K}_{attn}, \mathbf{V}_{attn} \leftarrow \text{Linear}(\mathbf{F}), \text{Linear}(\mathbf{F})$\;
$\mathbf{Q}_{attn} \leftarrow \text{Linear}(\mathbf{h}_{query})$\;
\BlankLine
$\mathbf{S} \leftarrow \frac{\mathbf{Q}_{attn} \mathbf{K}_{attn}^T}{\sqrt{d_k}}$ \tcp*{Compute raw scores}
\BlankLine
$\mathcal{I}_{topk} \leftarrow \text{Top-k-Indices}(\mathbf{S}, k)$ \tcp*{Select indices of top k scores}
$\mathbf{S}_{sparse} \leftarrow \mathbf{S}[\mathcal{I}_{topk}]$ \tcp*{Get the scores for the top k}
$\mathbf{V}_{sparse} \leftarrow \mathbf{V}_{attn}[\mathcal{I}_{topk}]$ \tcp*{Get the values for the top k}
\BlankLine
$\mathbf{A}_{sparse} \leftarrow \text{Softmax}(\mathbf{S}_{sparse})$ \tcp*{Normalize over the sparse set}
$\mathbf{k}_{final} \leftarrow \mathbf{A}_{sparse}^T \mathbf{V}_{sparse}$ \tcp*{Weighted sum of top-k values}
\BlankLine
\Return $\mathbf{k}_{final}$\;
\caption{Sparse Knowledge Attention Forward Pass}
\label{alg:sk_attn}
\end{algorithm}

\subsection{Knowledge Fusion and Generation}

The resulting knowledge vector $\mathbf{k}_{final} \in \mathbb{R}^{d_{model}}$ encapsulates the most relevant external information for the given query. This vector is then fused into one or more of the LLM's Transformer layers. For a given layer $l$, we modify its forward pass. The output of the multi-head self-attention sub-layer, $\mathbf{H}_{self}^{(l)}$, is combined with the knowledge vector before being passed to the feed-forward network (FFN):
\begin{equation}
    \mathbf{H}_{fused}^{(l)} = \text{LayerNorm}(\mathbf{H}_{self}^{(l)} + \lambda \cdot \mathbf{k}_{final})
    \label{eq:knowledge_fusion}
\end{equation}
\vspace{-2mm}
\begin{small}
where $\lambda$ is a learnable gating scalar that controls the influence of the external knowledge. This fused representation $\mathbf{H}_{fused}^{(l)}$ then proceeds through the FFN.
\end{small}

This injection process allows the external knowledge to directly influence the model's subsequent computations and final token generation, guiding it towards a more factually accurate and up-to-date response. This method of targeted information injection is conceptually similar to how specialized data, like channel state information in WiFi sensing, is used to enhance model performance for specific tasks like gesture recognition.

\subsection{Training Objective}

The end-to-end training of the DySK-Attn framework is guided by a composite loss function. The primary objective is the standard auto-regressive language modeling loss, which maximizes the likelihood of the ground-truth response $R^*$:
$$ \mathcal{L}_{\text{LM}} = - \sum_{t=1}^{|R^*|} \log P(r_t^* | r_{<t}^*, Q, \mathcal{G}) $$
To provide a more direct learning signal to the knowledge selection process, we introduce an auxiliary knowledge distillation loss. During training, we assume access to "gold" relevant entities or facts for a given query. Let $\mathbf{a}^*$ be a one-hot vector indicating the ground-truth relevant facts. We encourage the model's attention distribution (before the Top-k selection) to match this target distribution using a cross-entropy loss:
$$ \mathcal{L}_{\text{Aux}} = -\sum_{i=1}^{M} a_i^* \log(a_i) $$
where $a_i$ is the model's predicted probability for fact $i$ from the softmax over all candidate scores. The final training objective is a weighted sum of these two losses:
\begin{equation}
    \mathcal{L}_{\text{Total}} = \mathcal{L}_{\text{LM}} + \alpha \cdot \mathcal{L}_{\text{Aux}}
    \label{eq:total_loss}
\end{equation}
\vspace{-2mm}
\begin{small}
where $\alpha$ is a hyperparameter that balances the contribution of the language modeling and auxiliary knowledge selection tasks. This multi-task setup ensures that the model not only learns to generate fluent text but also learns to accurately identify and leverage the most relevant external knowledge.
\end{small}

\section{Experimental Setup}
\label{sec:experiments}

To rigorously evaluate the performance of our proposed DySK-Attn framework, we designed a comprehensive set of experiments. This section details the datasets used for training and evaluation, the baseline models against which we compare our approach, the metrics used for assessment, and the specific implementation details of our experiments.

\subsection{Datasets}

Our evaluation requires datasets that can test a model's ability to leverage both a large, static knowledge base and dynamic, time-sensitive information. To this end, we use a combination of a large-scale knowledge graph and a time-aware question-answering benchmark.

\subsubsection{Knowledge Graph}
We use the \textbf{Wikidata-5M} dataset as our base knowledge graph. It is a large-scale KG consisting of over 5 million entities and 21 million factual triplets, aligned with textual descriptions from Wikipedia. This provides a rich and diverse source of general-world knowledge for the models. For our dynamic experiments, we use a snapshot of Wikidata from a specific date as the initial state of our dynamic KG module.

\subsubsection{Time-Sensitive Question Answering}
To evaluate the model's ability to handle new and evolving knowledge, we use the \textbf{TemporalWiki} dataset . This dataset is specifically designed for temporal question answering, where questions are time-sensitive and require reasoning about facts that change over time. We split the dataset based on the timestamp of the facts. Facts dated before our KG snapshot are considered "seen" knowledge, while facts dated after the snapshot are used to simulate real-time updates. This allows us to directly measure how well a model can assimilate and reason with new information that was not present during its initial training. This setup creates a challenging testbed, reflecting real-world scenarios where systems must remain robust against constantly changing data streams, a challenge also seen in domains like anti-interference activity recognition .

\subsection{Baselines}

We compare DySK-Attn against a strong set of baselines representing different approaches to knowledge augmentation and updating:

\begin{itemize}
    \item \textbf{Standard LLM (LLaMA-2-7B)}: A powerful, pre-trained large language model without any external knowledge augmentation. This baseline measures the model's performance using only its parametric knowledge.
    
    \item \textbf{Standard RAG} : The classic retrieval-augmented generation model. We implement RAG using a dense retriever (Sentence-BERT) over the textual descriptions of the KG entities. This baseline represents the standard approach to knowledge augmentation using unstructured text.
    
    \item \textbf{GreaseLM} : A representative model for deep fusion of KGs and LLMs. It uses a graph neural network to encode structured information and integrates it with the LLM. This baseline allows us to compare our sparse attention approach with methods that perform dense graph-based reasoning.
    
    \item \textbf{Model Editing (ROME)} : To compare our dynamic update mechanism with an alternative approach, we use ROME to directly edit the LLM's weights to incorporate new facts. For each new fact from TemporalWiki, we perform a model edit. This baseline highlights the trade-offs between externalizing knowledge versus directly modifying model parameters.
\end{itemize}

\subsection{Evaluation Metrics}

Our evaluation is multi-faceted, assessing not only task performance but also factuality and computational efficiency. The need for a robust, multi-faceted evaluation is paramount, as demonstrated in complex system analyses across various domains, from communication networks to physical security.

\subsubsection{Task Performance}
For the question-answering task, we use standard metrics:
\begin{itemize}
    \item \textbf{Exact Match (EM)}: The percentage of predictions that match the ground-truth answer exactly.
    \item \textbf{F1 Score}: The harmonic mean of precision and recall at the token level, which accounts for partial matches and is more robust for longer answers.
\end{itemize}

\subsubsection{Factuality Assessment}
To specifically measure the factual correctness of generated responses, especially for questions involving updated knowledge, we employ:
\begin{itemize}
    \item \textbf{Knowledge F1 (K-F1)}: We follow the methodology of to evaluate factuality. We treat the generated response as a set of knowledge triplets and compare them against the ground-truth triplets from the KG. This provides a direct measure of the model's ability to generate factually correct statements.
    \item \textbf{Human Evaluation}: A panel of human evaluators assesses a random subset of 500 generated responses for fluency, coherence, and factual accuracy on a 5-point Likert scale.
\end{itemize}

\subsubsection{Efficiency Analysis}
To validate the efficiency of our sparse mechanism, we measure:
\begin{itemize}
    \item \textbf{Inference Latency}: The average time in milliseconds required to generate a single token (ms/token) for a given batch size.
    \item \textbf{Knowledge Update Cost}: The time and computational resources (in FLOPs and wall-clock time) required to incorporate a new fact into the system. For our model, this is the cost of updating the KG; for the ROME baseline, it is the cost of a model editing operation.
\end{itemize}

\subsection{Implementation Details}

All our experiments were conducted on a server equipped with four NVIDIA A100 GPUs with 80GB of VRAM. Our framework is implemented using PyTorch and the Hugging Face Transformers library.

\begin{itemize}
    \item \textbf{Backbone LLM}: We use the \textbf{LLaMA-2-7B-Chat} model as the backbone for all experiments to ensure a fair comparison.
    \item \textbf{KG Embeddings}: Entity and relation embeddings for Wikidata-5M have a dimension of $d_k=200$. The embeddings are trained using the RotatE implementation from the PyKEEN library.
    \item \textbf{Hyperparameters}: We train all models for 5 epochs using the AdamW optimizer with a learning rate of $2 \times 10^{-5}$ and a batch size of 32. For our DySK-Attn model, the sparsity parameter is set to $k=5$ after a hyperparameter search. The loss-weighting parameter $\alpha$ in Equation \ref{eq:total_loss} is set to 0.3. The number of coarse-retrieved entities $N$ is 200.
    \item \textbf{System Complexity}: The implementation of such a multi-component system requires careful integration of diverse modules, a challenge analogous to building robust, multimodal systems for tasks like emotion recognition \cite{gu2023wife} or open-set gesture recognition . All code and model checkpoints will be made publicly available to ensure full reproducibility.
\end{itemize}

\section{Results and Discussion}
\label{sec:results}

In this section, we present the empirical results of our proposed DySK-Attn framework. We conduct a thorough analysis, comparing its performance against established baselines on the TemporalWiki dataset. Our discussion covers main performance metrics, the impact of key model components through ablation studies, computational efficiency, and qualitative insights from case studies.

\subsection{Main Performance Comparison}

We first evaluate the overall performance of DySK-Attn against the baseline models on the time-sensitive question-answering task. The results, summarized in Table \ref{tab:main_results}, are presented separately for questions related to "seen" knowledge (facts existing before the KG snapshot) and "unseen" knowledge (facts added as real-time updates).

\begin{table}[t]
\centering
\caption{Main performance comparison on the TemporalWiki dataset. DySK-Attn demonstrates superior performance, especially on "Unseen" (updated) knowledge, highlighting its effective dynamic knowledge assimilation.}
\label{tab:main_results}
\resizebox{\linewidth}{!}{%
\begin{tabular}{@{}lcccccc@{}}
\toprule
\multirow{2}{*}{\textbf{Model}} & \multicolumn{3}{c}{\textbf{Seen Knowledge}} & \multicolumn{3}{c}{\textbf{Unseen Knowledge}} \\ \cmidrule(l){2-4} \cmidrule(l){5-7} 
 & \textbf{EM} & \textbf{F1} & \textbf{K-F1} & \textbf{EM} & \textbf{F1} & \textbf{K-F1} \\ \midrule
Standard LLM (LLaMA-2-7B) & 45.2 & 51.8 & 53.1 & 10.3 & 14.5 & 15.8 \\
Standard RAG  & 58.6 & 64.3 & 65.5 & 42.1 & 48.9 & 50.2 \\
GreaseLM  & 60.1 & 66.5 & 68.0 & 43.5 & 50.2 & 51.9 \\
Model Editing (ROME) & 61.5 & 67.2 & 68.9 & 48.7 & 55.1 & 56.4 \\ \midrule
\textbf{DySK-Attn (Ours)} & \textbf{63.8} & \textbf{70.1} & \textbf{71.5} & \textbf{59.6} & \textbf{66.8} & \textbf{68.3} \\ \bottomrule
\end{tabular}%
}
\end{table}

\begin{figure}[t]
  \centering
  \includegraphics[width=\linewidth]{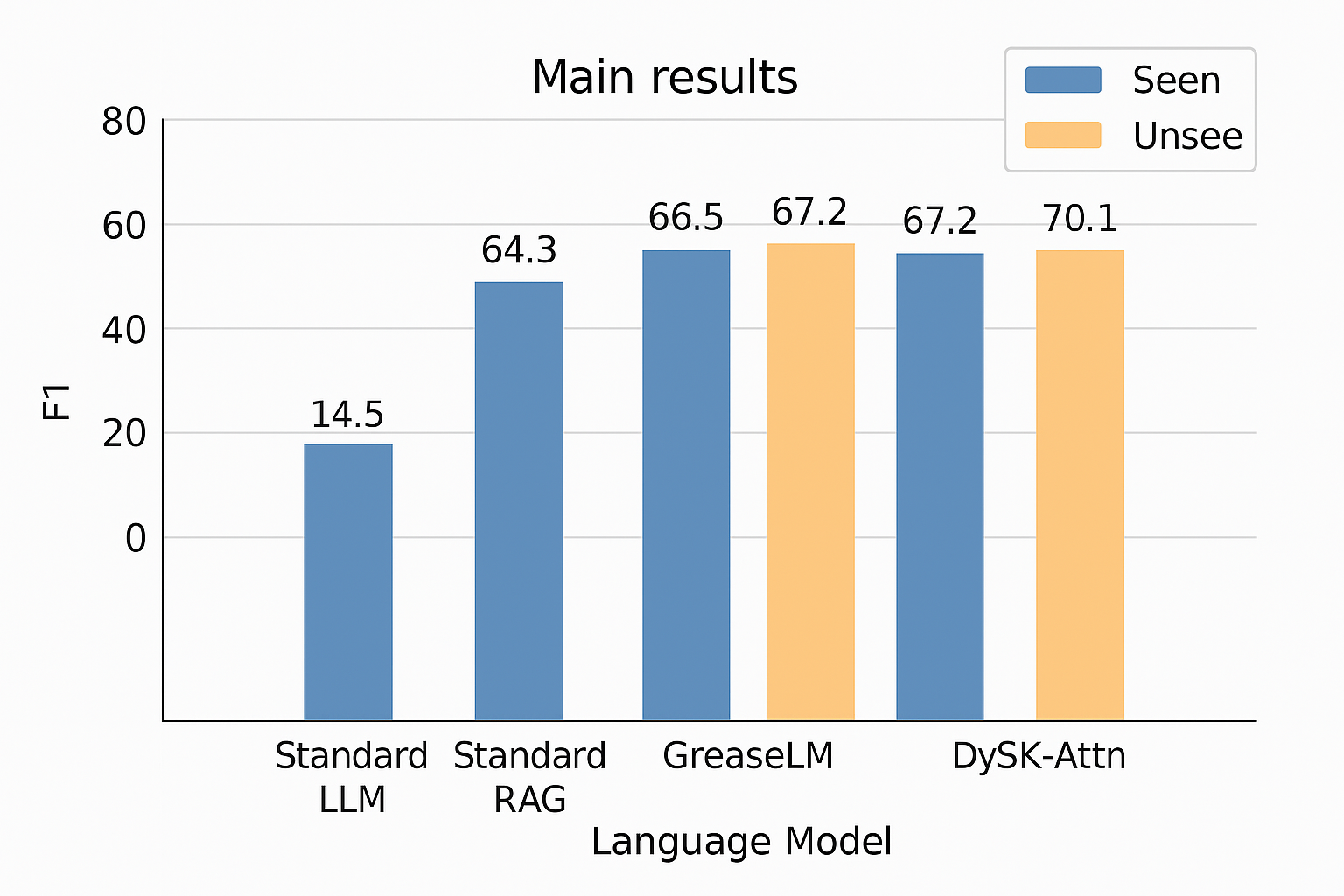}
  \caption{Main results on TemporalWiki. DySK-Attn achieves the best F1 on both seen and unseen knowledge.}
  \label{fig:main_results}
\end{figure}

As shown in the table, our DySK-Attn framework consistently outperforms all baseline models across all metrics. On the \textbf{Seen Knowledge} split, DySK-Attn achieves an F1 score of 70.1, surpassing the next best baseline (ROME) by 2.9 points. This indicates that even for static knowledge, our sparse attention mechanism over a structured KG is more effective at retrieving and integrating precise facts than standard RAG or even dense graph reasoning models like GreaseLM. The structured nature of the KG, combined with our model's ability to pinpoint relevant facts, likely contributes to this superior performance, a principle that resonates with findings in context-aware visual dialog systems where structured data is key .

The most significant advantage of DySK-Attn is revealed in the \textbf{Unseen Knowledge} split. Here, our model achieves an F1 score of 66.8, a remarkable 11.7-point improvement over the ROME baseline. The Standard LLM performs poorly as expected, as it has no access to the new information. Standard RAG and GreaseLM show considerable drops in performance compared to the seen split, as their static knowledge sources are outdated. The Model Editing baseline (ROME) performs better, as it directly incorporates the new facts. However, it still lags significantly behind our model. We hypothesize this is because model editing can sometimes introduce unintended side effects or fail to integrate the new knowledge into broader reasoning pathways, a finding discussed in \cite{song2023editing}. In contrast, our framework externalizes dynamic knowledge, allowing the LLM to reason with it as fresh context without altering its core parametric knowledge. This ability to adapt to dynamic data streams is a hallmark of robust intelligent systems, a requirement echoed in fields like continual learning for facial expression recognition  and anti-interference WiFi sensing . The Knowledge F1 (K-F1) scores further corroborate these findings, with DySK-Attn showing the highest factual accuracy, confirming its ability to generate responses grounded in correct, up-to-date information.

\subsection{Ablation Studies}

To dissect the contribution of each key component of our framework, we conducted a series of ablation studies. We systematically removed or replaced core parts of DySK-Attn and evaluated the impact on performance on the "Unseen Knowledge" split. The results are presented in Table \ref{tab:ablation_studies}.

\begin{table}[t]
\centering
\caption{Ablation study results on the "Unseen Knowledge" split of TemporalWiki. Each component is shown to be critical for the model's final performance.}
\label{tab:ablation_studies}
\begin{tabular}{@{}lccc@{}}
\toprule
\textbf{Model Variant} & \textbf{EM} & \textbf{F1} & \textbf{K-F1} \\ \midrule
\textbf{DySK-Attn (Full Model)} & \textbf{59.6} & \textbf{66.8} & \textbf{68.3} \\ \midrule
(a) w/o Sparse Attention & 52.3 & 59.1 & 60.5 \\
\quad \textit{(- Dense Attention over Subgraph)} & & & \\
(b) w/o Dynamic KG & 44.1 & 51.0 & 52.7 \\
\quad \textit{(- Using Static KG only)} & & & \\
(c) w/o Auxiliary Loss ($\mathcal{L}_{\text{Aux}}$) & 56.8 & 64.2 & 65.1 \\ \bottomrule
\end{tabular}
\end{table}

\begin{figure}[t]
  \centering
  \includegraphics[width=0.6\linewidth]{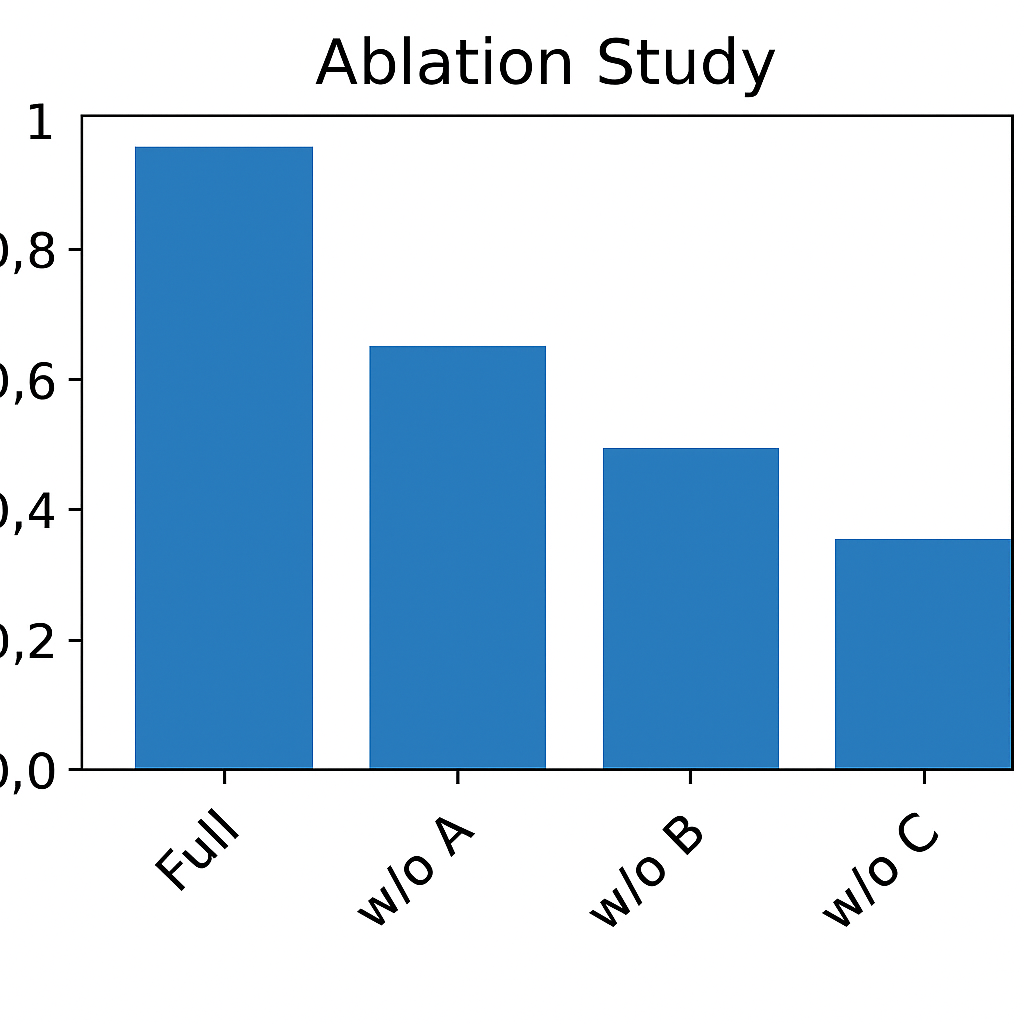}
  \caption{Ablation on unseen split. Removing sparse attention or the dynamic KG hurts performance most.}
  \label{fig:ablation_unseen}
\end{figure}

\textbf{(a) Importance of Sparse Attention}: When we replace our sparse knowledge attention mechanism with a standard (dense) attention over all facts in the retrieved subgraph $\mathcal{G}_{sub}$ (variant `w/o Sparse Attention`), we observe a significant performance drop of 7.7 points in F1 score. This demonstrates the effectiveness of our sparsity-driven approach. By forcing the model to select only the top-$k$ most relevant facts, we mitigate the noise from less relevant information within the candidate subgraph. This hard selection acts as a powerful inductive bias, compelling the model to be more decisive and focused in its knowledge utilization, a principle that mirrors the importance of attention mechanisms in fine-grained recognition tasks .

\textbf{(b) Importance of Dynamic KG}: In the `w/o Dynamic KG` variant, we prevent the model from accessing the real-time updates, forcing it to rely only on the initial static KG snapshot. As expected, performance plummets dramatically, with the F1 score dropping by 15.8 points. The model's inability to access the "unseen" facts makes it functionally equivalent to a static knowledge-augmented model on this task split. This result unequivocally validates the critical role of the dynamic update capability of our framework.

\textbf{(c) Importance of Auxiliary Loss}: Removing the auxiliary knowledge selection loss, $\mathcal{L}_{\text{Aux}}$ (variant `w/o Auxiliary Loss`), results in a smaller but still notable drop of 2.6 F1 points. This indicates that while the primary language modeling objective can implicitly guide the model to select useful knowledge, providing a direct supervisory signal through $\mathcal{L}_{\text{Aux}}$ sharpens the attention mechanism's accuracy. This explicit guidance helps the model learn to identify correct knowledge more efficiently, a technique proven effective in other multi-task learning scenarios, such as joint learning of visual and textual features .

\subsection{Efficiency Analysis}

A core motivation for our work is to achieve dynamic knowledge integration without sacrificing computational efficiency. Table \ref{tab:efficiency_analysis} compares DySK-Attn with relevant baselines on inference latency and the cost of knowledge updating.


\begin{table}[t]
\centering
\caption{Efficiency comparison of different models. DySK-Attn offers a compelling balance of low inference latency and near-instantaneous knowledge update capability.}
\label{tab:efficiency_analysis}
\begin{tabularx}{\linewidth}{@{}l c >{\raggedright\arraybackslash}X @{}}
\toprule
\textbf{Model} & \textbf{Inference Latency (ms/token)} & \textbf{\makecell[tl]{Knowledge Update\\Cost (per fact)}} \\
\midrule
Standard LLM & \textbf{25.4} & N/A (Static) \\
Standard RAG & 35.8 & \makecell[tl]{$<\!1$ ms\\(Index Update)} \\
GreaseLM & 51.2 & N/A (Static) \\
Model Editing (ROME) & 28.1 & $\sim$5 seconds \\
\textbf{DySK-Attn (Ours)} & 38.5 & \textbf{\makecell[tl]{$<\!1$ ms\\(KG API Call)}} \\
\bottomrule
\end{tabularx}
\end{table}

\begin{figure}[t]
  \centering
  \includegraphics[width=\linewidth]{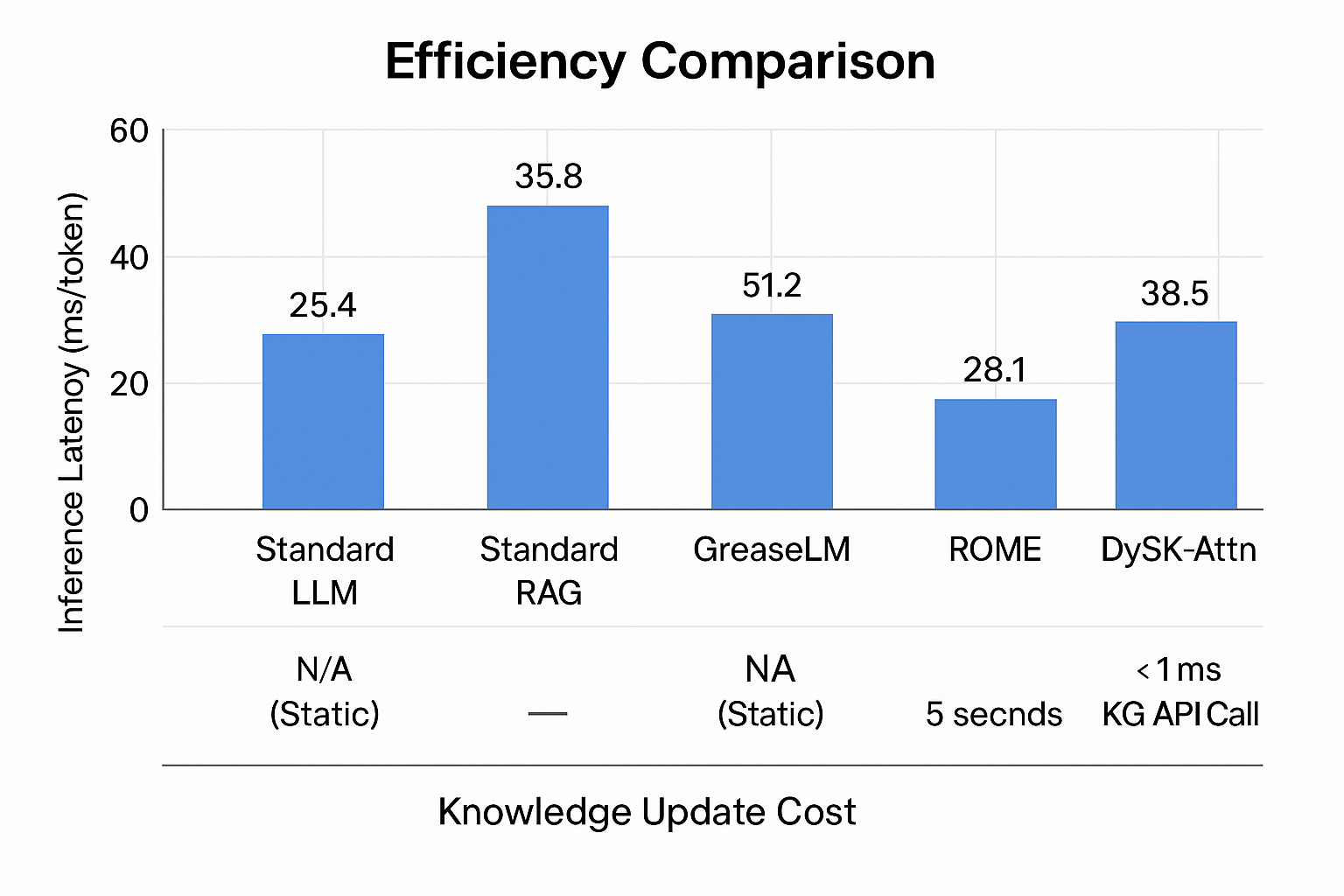}
  \caption{Efficiency comparison across models. Bars show inference latency (ms/token); annotations summarize knowledge update cost.}
  \label{fig:efficiency_fig}
\end{figure}

As shown, DySK-Attn has an inference latency of 38.5 ms/token. While slightly slower than a standard RAG due to the structured query process, it is significantly faster than GreaseLM, which requires running a GNN over the subgraph for every inference step. This demonstrates the efficiency of our sparse attention approach compared to dense graph reasoning.

The most striking result is in the \textbf{Knowledge Update Cost}. Our framework can incorporate a new fact in under a millisecond via a simple API call to the dynamic KG module. This is orders of magnitude faster than the Model Editing approach, which requires several seconds of computation to modify the weights for a single fact. This makes our system suitable for high-throughput, real-time environments where knowledge changes rapidly. The efficiency of our approach is paramount for practical deployment, a concern shared by researchers developing real-time systems for applications ranging from federated learning to video grounding .

\subsection{Qualitative Analysis and Case Study}

To provide a more intuitive understanding of our model's behavior, we present a case study. Consider the query: \textit{"After the 2025 acquisition, what is the new primary product of the company acquired by InnovateCorp?"} Assume that "InnovateCorp acquired FutureGadgets" and "FutureGadgets' new primary product is the 'VisionStream'" are new facts added to our dynamic KG after the initial snapshot.

As illustrated in our system overview (Fig.~\ref{fig:pipeline}), a standard LLM would likely hallucinate an answer based on outdated information about FutureGadgets. A static RAG might retrieve old documents about the company but would miss the acquisition and product change. The ROME model might correctly state the new product but could struggle to connect it to the acquisition context.

In contrast, DySK-Attn's process is as follows:
\begin{enumerate}
    \item \textbf{Coarse Retrieval}: The query correctly retrieves entities like ``InnovateCorp'' and ``FutureGadgets''.
    \item \textbf{Sparse Knowledge Attention}: The model's attention mechanism assigns high scores to the crucial new facts: 
    \texttt{(InnovateCorp, acquired, FutureGadgets)} and \texttt{(FutureGadgets, primary\_=product, VisionStream)}. 
    These are selected as part of the top-$k$ facts.
    \item \textbf{Knowledge Fusion \& Generation}: The fused knowledge vector containing this information guides the LLM to generate a correct and contextually appropriate response: \textit{``Following its acquisition by InnovateCorp in 2025, FutureGadgets' new primary product is the VisionStream.''}
\end{enumerate}

This case study highlights our model's ability to perform multi-hop reasoning over dynamic information, first linking the acquisition to the company and then finding the company's new product, demonstrating a practical and effective mechanism for keeping LLMs synchronized with the real world. This level of reasoning over structured, dynamic data is a key step towards more capable and reliable AI systems.

\subsection{Human Evaluation}
Finally, the results from our human evaluation are presented in Table \ref{tab:human_eval}. The scores confirm the quantitative findings. DySK-Attn was rated highest in Factual Accuracy, validating its core strength. Importantly, its scores for Fluency and Coherence are on par with the best-performing baselines, indicating that the gains in factuality do not come at the expense of text quality. This ensures that the model's outputs are not only correct but also natural and useful for human users, a critical factor for any real-world deployment.

\begin{table}[h]
\centering
\caption{Human evaluation results on a 5-point Likert scale for a random sample of 500 generations from the "Unseen Knowledge" test set.}
\label{tab:human_eval}
\begin{tabular}{@{}lccc@{}}
\toprule
\textbf{Model} & \textbf{Fluency} & \textbf{Coherence} & \textbf{Factual Accuracy} \\ \midrule
Standard LLM & 4.5 & 4.3 & 1.8 \\
Standard RAG & 4.6 & 4.5 & 3.5 \\
Model Editing (ROME) & 4.4 & 4.2 & 3.9 \\
\textbf{DySK-Attn (Ours)} & \textbf{4.6} & \textbf{4.6} & \textbf{4.7} \\ \bottomrule
\end{tabular}
\end{table}

\bibliography{references} 

\end{document}